\documentclass[10pt,twocolumn,letterpaper]{article}

\usepackage[pagenumbers]{cvpr} 

\definecolor{cvprblue}{rgb}{0.21,0.49,0.74}
\usepackage[pagebackref,breaklinks,colorlinks,allcolors=cvprblue]{hyperref}
\usepackage{mathrsfs}
\usepackage{adjustbox}
\usepackage{multirow}
\usepackage{multirow}
\usepackage{multicol}
\usepackage{tcolorbox}
\usepackage{changepage}
\usepackage{graphicx}
\newcommand{\VarSty}[1]{\textnormal{\ttfamily\color{blue!90!black}#1}\unskip}

\def\paperID{5209} 
\def\confName{CVPR}
\def\confYear{2025}

\title{Prompt-A-Video: \\Prompt Your Video Diffusion Model via Preference-Aligned LLM}

\def\authorBlock{
    Yatai Ji$^\textnormal{1}$\footnotemark[1] \quad
    Jiacheng Zhang$^\textnormal{1}$\footnotemark[1] \quad
    Jie Wu$^\textnormal{2}$\footnotemark[2] \quad
    Shilong Zhang$^\textnormal{1}$ \quad
    Shoufa Chen$^\textnormal{1}$ \quad  
    Chongjian Ge$^\textnormal{1}$ \\
    Peize Sun$^\textnormal{1}$ \quad
    Weifeng Chen$^\textnormal{2}$ \quad
    Wenqi Shao$^\textnormal{3}$ \quad
    Xuefeng Xiao$^\textnormal{2}$ \quad
    Weilin Huang$^\textnormal{2}$ \quad
    Ping Luo$^\textnormal{1}$\footnotemark[3]
    \and
    $^\textnormal{1}$The University of Hong Kong \qquad
    $^\textnormal{2}$ByteDance \qquad
    $^\textnormal{3}$Shanghai AI Lab
}
\author{\authorBlock}

\begin{document}

\twocolumn[{%
\renewcommand\twocolumn[1][]{#1}%
\maketitle
\begin{center}
    \vspace{-20pt}
    \includegraphics[width=0.99\linewidth]{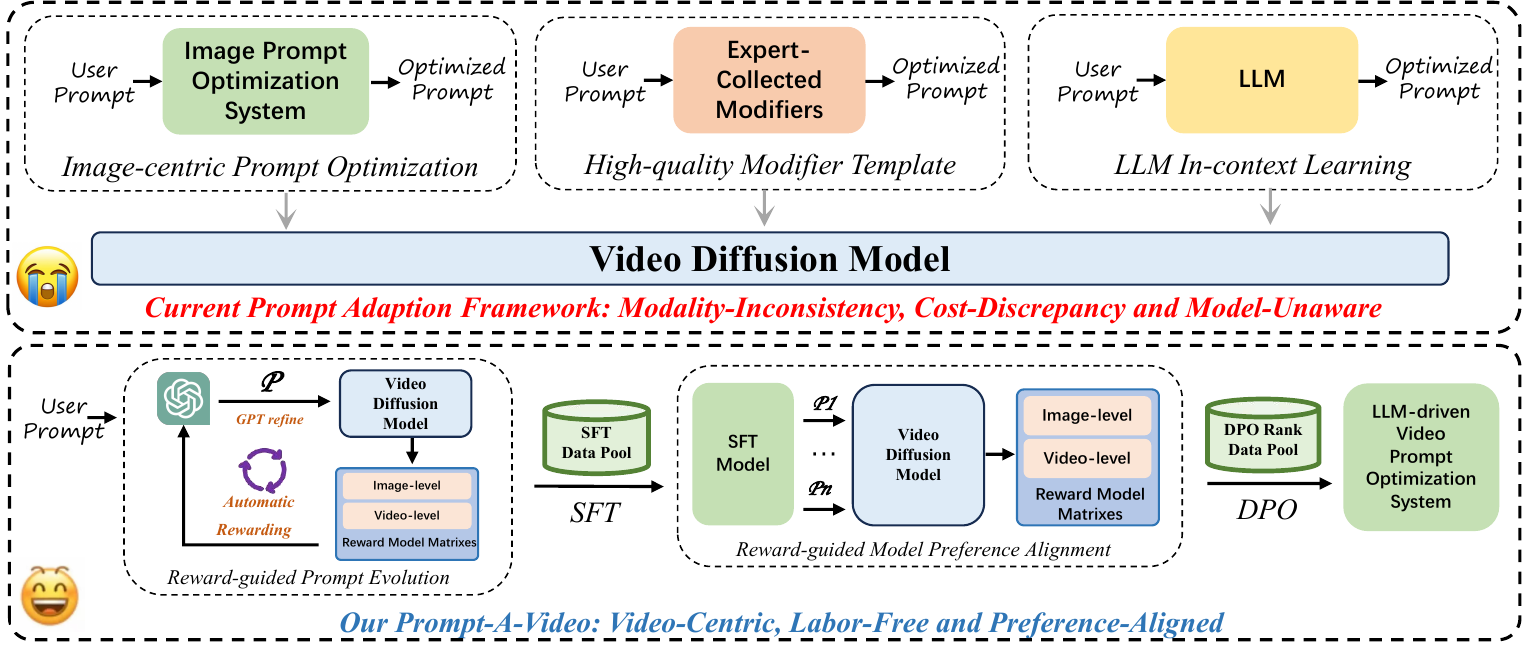}
\captionof{figure}{Illustration of the discrepancies between current prompt adaption frameworks and our proposed LLM-driven video prompt optimization system, namely \textbf{\textit{Prompt-A-Video}}. 
Current prompt adaption frameworks predominantly focus on prompt systems tailored for images, collecting intrinsic modifier templates, or rely on the in-context capabilities of LLM. However, in the realm of text-to-video, these methods encounter challenges stemming from \textcolor{red}{Modality-Inconsistency, Cost-Discrepancy, and Model-Unaware}. In this paper, we introduce Prompt-A-Video, a two-stage optimization and alignment system based on AI feedback, which aims to provide \textcolor{blue}{Video-Centric, Labor-Free and Preference-Aligned} prompts.
}
    \vspace{3pt}
    \label{fig:teaser}
    \end{center}%
}]
{
  \renewcommand{\thefootnote}%
    {\fnsymbol{footnote}}
  \footnotetext[1]{Co-first authors.}
  \footnotetext[2]{Project lead.}
  \footnotetext[3]{Corresponding Author.}
}


\begin{abstract}

Text-to-video models have made remarkable advancements through optimization on high-quality text-video pairs, where the textual prompts play a pivotal role in determining quality of output videos. 
However, achieving the desired output often entails multiple revisions and iterative inference to refine user-provided prompts. 
Current automatic methods for refining prompts encounter challenges such as Modality-Inconsistency, Cost-Discrepancy, and Model-Unaware when applied to text-to-video diffusion models. 
To address these problem, we introduce an LLM-based prompt adaptation framework, termed as \textbf{Prompt-A-Video}, which excels in crafting Video-Centric, Labor-Free and Preference-Aligned prompts tailored to specific video diffusion model.
Our approach involves a meticulously crafted two-stage optimization and alignment system. 
Initially, we conduct a reward-guided prompt evolution pipeline to automatically create optimal prompts pool and leverage them for supervised fine-tuning (SFT) of the LLM. 
Then multi-dimensional rewards are employed to generate pairwise data for the SFT model, followed by the direct preference optimization (DPO) algorithm to further facilitate preference alignment. 
Through extensive experimentation and comparative analyses, we validate the effectiveness of Prompt-A-Video across diverse generation models, highlighting its potential to push the boundaries of video generation.

\end{abstract}    
\section{Introduction}
\label{sec:intro}

Text-to-Video generation~\cite{cogvideo, sora, easyanimate, DBLP:journals/corr/abs-2305-13840/controlavideo, DBLP:conf/iclr/VillegasBKM0SCK23/phenaki, ho2022video/VDM} has witnessed remarkable progress in recent years, prominently driven by large-scale high-quality text-video pairs. 
In text-to-video generation frameworks, text serves as the exclusive conditioning signal, playing a decisive role in determining the output videos' content and quality. 
However, empirical evidence highlights that standard user inputs often fail to prompt the creation of aesthetically appealing videos. This deficiency stems from the fact that contemporary high-fidelity video generation models, such as Open-Sora~\cite{opensora} and CogVideoX~\cite{cogvideox}, predominantly utilize intricate descriptions generated by large visual language models (LVLM)~\cite{DBLP:journals/corr/abs-2407-07577/idavlm, DBLP:conf/nips/LiuLWL23a/llava, DBLP:journals/corr/abs-2308-12966/qwenvl, DBLP:journals/corr/abs-2404-16994/pllava, DBLP:journals/corr/abs-2306-07096/glscl, DBLP:conf/cvpr/JiTJKCZWY023/scl, DBLP:journals/corr/abs-2312-14238/internvl} as training prompts, far exceeding the quality of human-written inputs. These detailed prompts possess enhanced expressiveness and complexity, thus we need to recalibrate the input prompts to align with the model's preference for optimal performance.

Previous researches have extensively explored prompt engineering for improving visual generation, encompassing various approaches: 
utilizing high-quality modifier templates~\cite{DBLP:conf/chi/LiuC22a/guideprompt, oppenlaender2023taxonomy/promptmodifier} from experts and open-source communities, 
developing text-to-image prompt optimization systems~\cite{promptist, dynamic_boost, pivot_boost}, learning to append modifiers related to technical specifications and artistic styles, 
and leveraging GPT's in-context learning ability to expand user prompts~\cite{cogvideox, opensora}. 
However, direct application of these methodologies to text-to-video generation reveals significant limitations:

\begin{itemize}[leftmargin=*]
\item \textit{\textbf{Modality-Inconsistency}}
Image prompts predominantly accentuate static attributes like composition and color, inadequately addressing the dynamic requisites of video creation, like motion fluency, narrative coherence, and scene transitions. 
Moreover, in contrast to image prompt enhancement techniques that rely on appending quality modifiers, text-to-video models are less responsive to overall quality descriptors.

\item \textit{\textbf{Cost-Discrepancy.}}
There are several established platforms offering high-quality prompts for text-to-image generation, but video models lag behind image counterparts in maturity.
To obtain the optimal prompt for specific text-to-video model, it requires extensive domain expertise and and community's exploration.


\item \textit{\textbf{Model-Unaware.}}
Recent attempts in video prompt engineering expand and enrich user prompts based on in-context learning ability of GPT. However, the refined prompts do not adequately consider video quality and lack alignment with specific video generation models.

\end{itemize}

Consequently, a pertinent query emerges:   \textit{How can we devise a model-aware system to facilitate generating video-centric, labor-free, and preference-aligned prompts?}



In this paper, we introduce Prompt-A-Video, an LLM-driven automatic prompt adaptation framework meticulously crafted for enhancing text-to-video models.
As shown in Figure \ref{fig:teaser}, this framework unfolds in two distinct stages, each characterized by individual data preparation and model optimization pipelines. 
We first establish a comprehensive, multi-dimensional reward system tailored to assess the quality of generation results from diverse prompts while ensuring fidelity to the original user intentions. 
In the initial phase, to avoid labor-intensive manual engineering, we design a prompt refinement pipeline, inspired by evolutionary algorithms~\cite{DBLP:conf/iclr/Guo0GLS0L0Y24/EA1, DBLP:journals/tec/ZhangS09/EA3, DBLP:journals/jgo/StornP97/EA4}, which iteratively produce model-preferred prompts. 
The pipeline, namely reward-guided prompt evolution, uses LLM as evolutionary operator to synthesize and refine historical prompts, with selection guided by reward models. 
This evolutionary pipeline serves dual purposes: it not only optimizes video generation quality through iterative prompt refinement, but also creates training data pairs of original and refined prompts. 
Leveraging this curated dataset, we fine-tune the LLM, endowing it with foundational capabilities for prompt enhancement. 
The subsequent phase involves employing the reward model to rank candidates generated by the Supervised Fine-Tuning~(SFT) model. Employing Direct Preference Optimization~(DPO)~\cite{DBLP:conf/nips/RafailovSMMEF23/dpo}, we augment the alignment of generated prompts with the inherent preference of the video generation model.\protect\footnotemark[1]

The contributions of this work can be summarized as:
\begin{itemize}
\item ~\underline{\textit{New Insights}}:
To the best of our knowledge, we offer the first attempt to design a \textbf{video-centric} prompt optimization system. Prompt-A-Video liberates the text-to-video model from the constraint of meticulously crafted user inputs, paving a new way to enhancing video generation quality.

\item ~\underline{\textit{Automatic Data Pipeline}}:
Leveraging multi-dimensional rewards and reward-guided prompt evolution pipeline, we achieve \textbf{labor-free} generation of model-specific, high-quality prompt datasets for diverse text-to-video models. 

\item ~\underline{\textit{Optimization and Alignment}}:
The integration of Supervised Fine-Tuning (SFT) optimization techniques and Direct Preference Optimization (DPO) alignment strategies reinforces Prompt-A-Video's efficacy in prompt enhancement for specific generation model, deriving a \textbf{preference-aligned} LLM.

\item ~\underline{\textit{Excellent performance and generalization}}:
Extensive evaluations reveals that Prompt-A-Video's superiority on enhancing video results across various evaluation metrics. Concurrently, significant improvements are achieved when applying Prompt-A-Video to text-to-image scenarios, underscoring its robustness and generalizability.
\end{itemize}

\footnotetext[1]{Our codes are available at \url{https://github.com/jiyt17/Prompt-A-Video}.}
\section{Related Works}
\label{sec:relatedwork}

\subsection{Diffusion-based Video Generation}
Diffusion models have exhibited remarkable performance in generating high - quality videos. Early efforts ~\cite{animatediff,magicvideo,lavie,modelscope,alignvideo,tuneavideo,makeavideo,text2video} typically expand the pre-trained image diffusion model ~\cite{ldm} into the video diffusion model by integrating temporal layers (e.g., temporal convolution or temporal attention). For instance, Animatediff ~\cite{animatediff} dissociates video generation into content and motion generation and appends an additional motion module to the base image generation model to accomplish video clip generation. Subsequently, large video pre-trained diffusion models ~\cite{videocrafter1,videocrafter2,svd,lumiere}, with VideoCrafter ~\cite{videocrafter1} as a representative work, display impressive video quality thanks to the large-scale video pre-training dataset. However, these methods encounter limitations in generating long videos due to the inherent restrictions on capacity and scalability within the UNet design.
Pioneered by seminal works such as Sora ~\cite{sora}, a succession of DiT - based video diffusion models ~\cite{easyanimate, opensora,vidu,cogvideo,cogvideox} has emerged steadily. Leveraging the large-scale training and the scalability of the DiT architecture, these models are capable of generating longer videos of up to one minute. Despite these progresses, the current text-to-video (T2V) generation remains sensitive to the input prompt, thereby posing a challenge for users to input an appropriate one. In this study, our objective is to develop an effective prompt enhancement system for video generation, which can substantially alleviate user attempt cost.
\subsection{Image Prompt Refinement}

Prompt adaptation system is aimed at improving the generation quality (such as visual quality, aesthetic allure, text - image alignment, etc.) of prompt-based generative models ~\cite{gpt4,llam3,touvron2023llama,imagen,ldm} without altering the user's original intention by refining the input prompt. A plethora of works ~\cite{lin2024prompt,sabbatella2024prompt,ma2024large} have been proposed to explore prompt boost techniques in the domain of LLM, inspiring research into such techniques in the field of text-to-image generation ~\cite{zhong2023adapter,promptist,wu2024universal,dynamic_boost,pivot_boost,consitency_boost,wang2024discrete}.
Among these works, Promptist ~\cite{promptist} utilizes reinforcement learning (i.e., PPO) to refine the user's initial prompt based on the aesthetic and text-image alignment rewards of the generated image, making it more suitable for the diffusion model. \citet{dynamic_boost} propose the prompt auto-edit method (PAE) to enhance the initial user prompt. This method consists of a fine-tuned and reinforcement learning stage to dynamically fine-control the weights and injection time steps of each word within a prompt. \citet{pivot_boost} develop prompt refinement with an image pivot. It views the image as a bridge and converts the data-scarce prompt refinement into a data-rich process, ultimately resulting in a superior prompt refinement model trained with abundant data. More recently, ~\cite{consitency_boost} focused on text-image consistency and designed a tuning-free prompt optimization system by leveraging the LLM to iteratively generate revised prompts that maximize the consistency score.
Despite the efficacy of these methods, they seldom explore prompt refinement for video generation. In this paper, we propose a prompt adaptation system customized for text-to-video generation.
\section{Method}
\label{sec:method}
In this section, we present Prompt-A-Video, an automatic LLM-based prompt adaptation framework designed for video generation. Our primary objective is to improve the quality of generated videos by leveraging a Large Language Model (LLM) that transforms concise user inputs into rich, detailed prompts tailored for specific generation models. We first introduce our multi-dimensional reward system in Section \ref{sec: reward}, which provides a comprehensive video-centric evaluation tool for subsequent methods. To address the challenge of limited high-quality video prompt datasets, Section \ref{sec:EPR} outlines our reward-guided prompt evolution pipeline, serving as a labor-free data engine to produce model-preferred prompts automatically. Section \ref{sec:boost} details our two-stage optimization approach: first fine-tuning the LLM with these curated data pairs, followed by Direct Preference Optimization (DPO) to align with preference of generation models. 


\begin{figure*}[t]
	\centering
	\includegraphics[width=\linewidth]{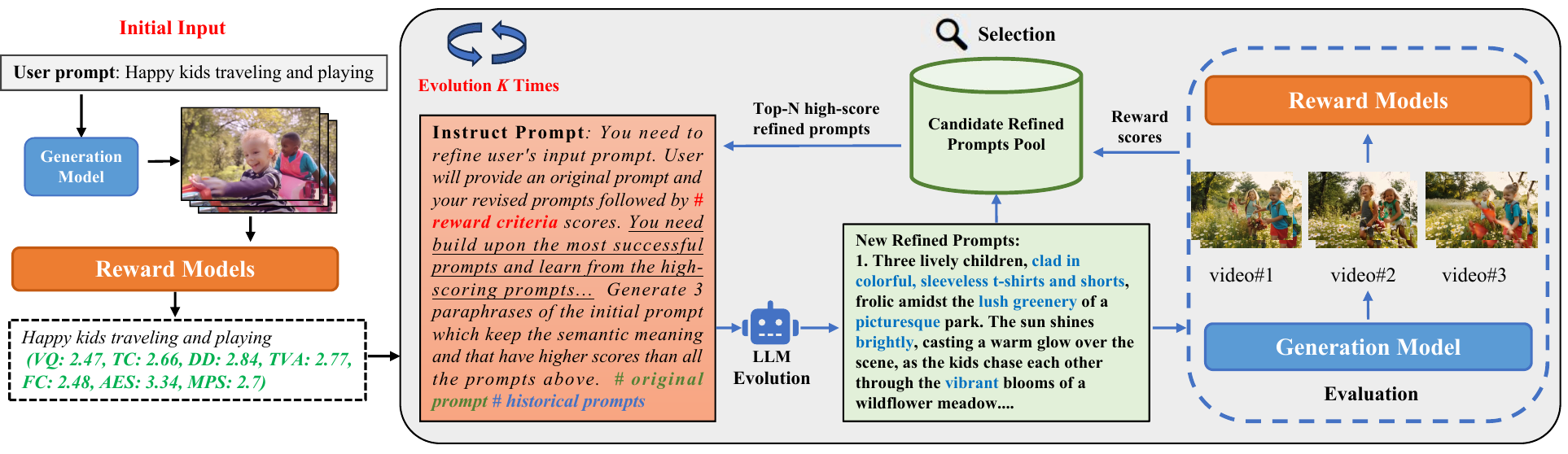}
	\caption{The Pipeline of Reward-guided Prompt Evolution, which employs iterative reward-feedback loops to obtain superior prompts through three processes: evaluation, selection, and evolution. This evolutionary pipeline serves dual purposes: it not only optimizes video generation quality through iterative prompt refinement, but also creates training data pairs of original and refined prompts. }
	\label{fig:pipeline}
 \vspace{-0.2cm}
\end{figure*}

\subsection{Multi-dimensional Reward System}
\label{sec: reward} 
To thoroughly assess the efficacy of our optimized prompts, we employ a range of specialized reward models to evaluate the generated videos. Our reward system is meticulously designed from two perspectives: image-level and video-level.

Initially, we adopt image-level rewards to measure the quality of video frames. To quantify aesthetic attributes of the images, we utilize Aesthetic Predictor\protect\footnotemark[1] built upon a CLIP~\cite{DBLP:conf/icml/RadfordKHRGASAM21/clip} encoder, like in the previous text-to-image prompt boost methods. Additionally, we incorporate the Multi-dimensional Human Preference Score (MPS)~\cite{MPS}, a competitive text-to-image reward model, to assess the overall preference for the images. 

\footnotetext[1]{https://github.com/LAION-AI/aesthetic-predictor}

For video-level evaluation, we adopt VideoScore~\cite{DBLP:conf/emnlp/HeJZKSSCCJAWDNL24/videoscore}, a comprehensive Large Vision Language Model that evaluates generated videos across five critical dimensions: visual quality (VQ), temporal consistency (TC), dynamic degree (DD), text-to-video alignment (TVA), and factual consistency (FC). VQ appraises videos' quality in terms of clearness, resolution, brightness, and color. TC, a distinctive feature of videos compared to static images, measures the consistency of objects or humans across frames, crucial for maintaining a coherent video. DD quantifies the degree of dynamic changes within the video. TVA ensures the semantic alignment between the video content and the original input prompt, preserving the user's intention. FC assesses the video's adherence to common-sense and factual knowledge, verifying the absence of artifacts. The video-level rewards facilitate identifying prompts that are conducive to high-quality video generation.

\subsection{Reward-guided Prompt Evolution}
\label{sec:EPR}


Previous prompt boost methods for image generation always collect descriptions with suffix modifiers as fine-tuning data, like Lexica.art\protect\footnotemark[1]. They always only append modifier words to enhance the art style or emphasize the image quality. However, due to the modality-inconsistency, text-to-video models are less responsive to general quality descriptions and require more expressive and detailed prompts. Therefore, we develop an automatic pipeline to refine preference-aligned prompts for the specific generation model, drawing inspiration from evolutionary algorithms (EAs) known for their performance and fast convergence.

\footnotetext[1]{https://lexica.art/}

Evolutionary algorithms typically follow these steps: an initial population of solutions is evaluated to determine fitness scores. Based on these scores, individuals are selected for operations such as crossover and mutation to create new solutions. These new solutions are also scored, and those with higher fitness in the entire population are selected for the next iteration. The process repeats until a predetermined score or maximum number of cycles is reached.

As illustrated in Figure \ref{fig:pipeline}, based on the idea of EAs, we propose reward-guided prompt evolution pipeline, which uses GPT-4o~\cite{gpt4} as the evolutionary operator to generate model-preferred prompts. Past researches~\cite{DBLP:conf/iclr/Guo0GLS0L0Y24/EA1, DBLP:journals/corr/abs-2206-08896/EA2} showed that LLMs have the ability to mimic operations in EAs. In our data construction pipeline, the initial population is the original user prompt. We define prompts' fitness scores that evaluate the video generated by the prompt with our multi-dimensional reward system. Specifically, the algorithm initializes with the input prompt and proceeds as follows. 

\begin{itemize}

\item \textit{Evaluation}: For the input prompt or newly generated prompts, we first create the corresponding videos. Then we score them with our multi-dimensional reward system, which has seven metrics in total. Finally, we append the scores behind the corresponding prompts, rendering candidates for evolution.

\item \textit{Selection}: In the second and subsequent iterations, we select the most effective prompts based on their scores. By merging new prompts with the existing population, we consider all metrics comprehensively to select top N prompts. 

\item \textit{Evolution}: We input the initial and selected prompts into GPT-4o to produce refined prompts. We design instructions for GPT-4o to incorporate historical experiences and combine different input prompts according to their scores. Moreover, we provide high-quality video descriptions as examples. To improve the efficiency of optimal prompt searching, GPT-4o outputs three refined prompts at once. Then we return to step 1. The specific GPT-4o instructions are detailed in the Appendix A.

\end{itemize}

We set a maximum number of iterations. As the process continues, the overall scores of selected refined prompts improve. At the end of the iterations, we select a target refined prompt whose scores in each dimension exceed a predefined threshold and have the highest overall score among all prompts. This allows us to construct data pairs of original and target prompts, with the target prompt favored by the specific generation model. 

\subsection{Two-stage Optimization}
\label{sec:boost}
To achieve automatic prompts refinement, we develop a prompt optimizer built upon LLama3-Instruct~\cite{DBLP:journals/corr/abs-2407-21783/llama3} as the foundation model, employing a two-stage training strategy. 
As shown in the lower part of Figure \ref{fig:teaser}, we first leverage supervised fine-tuning (SFT) to adapt the LLM for video-centric prompt enhancement, then use Direct Preference Optimization (DPO) to align the prompts with generative videos' preferences.

The curated SFT dataset comprises paired prompts $\{ (x, y) \}$, where $x$ represents the source prompt and $y$ denotes the corresponding target prompt derived from evolutionary data engine. To preserve the LLM's basic language capabilities while adapting it for prompt refinement, we finetune the model with LoRA-based approach. Specifically, we design task instruction $s$ and train with language modeling objective as follows:
\begin{align}
\mathcal{L}_{\text{SFT}} = - \mathbb{E}_{(x, y)} \operatorname{log} p(y | s, x).
\end{align}
In this way, the LLM has the instruction following ability for prompt refinement.

Following the supervised fine-tuning phase, we employ DPO, a more stable alternative to proximal policy optimization (PPO), to further enhance the LLM. DPO focuses on fine-tuning models by directly optimizing for user preferences. It enhances model performance by adjusting outputs to align with preferred outcomes, based on feedback or reward signals. 

When constructing DPO data, for each input prompt, we utilize the SFT model to generate five distinct refined versions, which are used to produce videos. Then these generated videos undergo comprehensive evaluation using our reward models, including Videoscore, Aesthetic Predictor and MPS. The evaluation metrics are summed after being normalized to the same scale. Finally, we identify the best and worst prompts based on their corresponding video scores, creating \{$prompt, chosen, rejected$\} triplets for training. This curated dataset then serves for optimizing the SFT model with the DPO objective function:
\begin{equation}
\footnotesize
\begin{split}
&\mathcal{L}_\text{DPO}(\pi_{\theta}; \pi_{ref}) =\\ 
&-\mathbb{E}_{(x, y_w, y_l)\sim \mathcal{D}}\left[\log \sigma \left(\beta \log \frac{\pi_{\theta}(y_w\mid x)}{\pi_{ref}(y_w\mid x)} - \beta \log \frac{\pi_{\theta}(y_l\mid x)}{\pi_{ref}(y_l\mid x)}\right)\right],
\end{split}
\label{eq:optimum_model}
\end{equation}
where $x$ donates the original prompt, and $y_w$, $y_l$ serve as the chosen and rejected prompts, respectively. 
To ensure continuous improvement, we execute multiple iterations of the DPO process, generating new triplet data based on the model trained with DPO from last iteration.

\section{Experiments}
\label{exp}

\subsection{Experimental Setup}

\noindent\textbf{Data Collection}. In the two-stage training process, the WebVid dataset serves as the source of video prompts.
Our reward-guided prompt evolution pipeline undergoes 4 iterative cycles, with 3 new prompts being generated per iteration and top 3 prompts being selected for next iteration. For each generation model, approximately 3,000 prompt pairs are collected. 
During the DPO phase, approximately 2,000 prompt triples are used for preference optimization. Based on our exploratory experiments, we conduct two iterations of DPO, as additional DPO iterations does not yield further improvements.

For our experimental evaluation, in accordance with previous research, we evaluate the effectiveness of prompt enhancement across both in - domain and out-of-domain test sets. For the in-domain evaluation, 130 prompts are randomly sampled from the WebVid dataset, and the VideoScore metric is utilized to measure the results. The out-of-domain evaluation makes use of the public benchmark VBench. Given that VBench encompasses a large number of prompts, many of which share the same main content but differ in style, we meticulously curate a representative subset of prompts. Prompts are selected from four dimensions of VBench: human action, overall consistency, appearance style, and subject consistency. Priority is given to prompts containing dynamic actions rather than static descriptions, resulting in a challenging test set of 100 prompts. VBench provides a comprehensive evaluation across six dimensions for the generated videos: subject consistency, background consistency, aesthetic quality, imaging quality, motion smoothness, and dynamic degree.

\noindent\textbf{Implementation Details}.
To verify the effectiveness of our Prompt-A-Video, extensive experiments are conducted across a variety of visual generation models. For the video generation task, several representative text-to-video models based on the Diffusion Transformer architecture are employed, such as Open-Sora 1.2~\cite{opensora} and CogVideoX~\cite{cogvideox}.
Diverse video generation configurations are implemented to showcase the versatility of our approach. Specifically, Open-Sora 1.2 is utilized to generate 2-second videos with a resolution of $720\times 1080$, while CogVideoX is employed for generating 4-second videos at  $480\times 720$ resolution.
The detailed training settings for the two-stage optimization are shown in Appendix B.

\noindent\textbf{Comparison Baselines}. In text-to-video generation, our refined prompts are compared against three types of prompts: (i) original prompts from the test set, (ii) prompts directly generated by a text-to-image prompt boost model~\cite{promptist}, and (iii) prompts refined by state-of-the-art large language models, such as GPT-4o~\cite{gpt4} (used by Open-Sora) and GLM-4~\cite{DBLP:journals/corr/abs-2406-12793/glm4} (employed by CogVideoX).

\subsection{Evaluation on Text-to-video}

\begin{table*}[]
	\centering
	\begin{adjustbox}{max width=\textwidth}
		\begin{tabular}{lcccccccccccc}
			\toprule[1pt]
			\multicolumn{1}{l|}{\multirow{2}{*}{\textbf{Model}}} & \multicolumn{6}{c|}{\textbf{WebVid (In-Domain)}}                                                                                                                             & \multicolumn{6}{c}{\textbf{VBench (Out-of-Domain)}}                                                                                                                 \\
			\multicolumn{1}{l|}{}                       & VQ           & TC           & DD & TVA &  FC &\multicolumn{1}{c|}{Avg}       & SC           & BC           & AQ & IQ &  MS & DD \\ \midrule
			\multicolumn{12}{l}{\textit{Open-Sora 1.2}}                                                                                                                                                                                                                                                                                                    \\ \midrule
			\multicolumn{1}{l|}{Original prompts}                  & 3.079 & 3.084 &  \underline{3.203}  & 3.156 &  3.060                & \multicolumn{1}{c|}{3.116}                     & 0.953           & 0.965           & 0.522                      &  \underline{0.630}                     & 0.987                     &  \textbf{0.44}                       \\
                \multicolumn{1}{l|}{Promptist}                  &   2.959 & 2.973 & 2.977 & 2.955 & 2.930     & \multicolumn{1}{c|}{2.959}                     &     0.962 & \underline{0.971} &  0.533 &  0.617 & 0.991 & 0.28 \\
                \multicolumn{1}{l|}{GPT-4o}                  & 3.014 & 3.082 & 3.103 &  3.167 & 3.031      & \multicolumn{1}{c|}{3.079}                     & \underline{0.964}        & \underline{0.971}       & 0.546                   & 0.603                   & \underline{0.992}              & \underline{0.37}                    \\
                \multicolumn{1}{l|}{Prompt-A-Video (SFT)}                  & 3.029 & 3.077 & 3.113 & 3.152 & 3.044 & \multicolumn{1}{c|}{3.083}                     & 0.959      &  0.970         &  0.531                 &  0.588                &  \underline{0.992}                 &  0.29                      \\
                \multicolumn{1}{l|}{Prompt-A-Video (DPO-1)}                  & \underline{3.090} &  \underline{3.152} &  3.198 & \underline{3.180} & \underline{3.108}   & \multicolumn{1}{c|}{\underline{3.146}}                     &   \textbf{0.968}       &   \textbf{0.972}          &   \underline{0.555}                      &  0.621                &   \textbf{0.993}                   & 0.26                       \\
                \multicolumn{1}{l|}{Prompt-A-Video (DPO-2)}                  & \textbf{3.254} &  \textbf{3.286} &  \textbf{3.411} & \textbf{3.358} & \textbf{3.282}   & \multicolumn{1}{c|}{\textbf{3.318}}                     &   0.962       &   0.970          &   \textbf{0.558}                      &  \textbf{0.658}                &   0.990                   & 0.29                       \\
			 \midrule
			\multicolumn{12}{l}{\textit{CogVideoX}}                                                                                                                                                                   \\ \midrule
			\multicolumn{1}{l|}{Original prompts}                  & 2.899 & 2.886 &  \textbf{3.186} & 3.167 & 2.808  & \multicolumn{1}{c|}{2.989}                     & 0.946  &  0.959 &   0.545 
             & 0.636  &  0.976  &   \textbf{0.70}                 \\
                \multicolumn{1}{l|}{Promptist}                  &     2.800 & 2.826 & 3.016 & 3.005 & 2.719   & \multicolumn{1}{c|}{2.873}                     &        \textbf{0.958} &  \textbf{0.966} & 0.561 & 0.632 &  \textbf{0.984} & 0.56        \\
                \multicolumn{1}{l|}{GLM-4}                  & 2.878 & 2.948 & 3.139 & 3.184 & 2.833    & \multicolumn{1}{c|}{2.996}                     &     0.951  &  \underline{0.963}  &  0.614  & 0.647  &  0.982  &  \underline{0.60}                \\
                \multicolumn{1}{l|}{Prompt-A-Video (SFT)}                  & 2.77 & 2.842 & 3.045 & 3.071 & 2.706 & \multicolumn{1}{c|}{2.887}                     & 0.935  &  0.954 &  0.590 &  0.617  &  0.978  & \textbf{0.70}                     \\
                \multicolumn{1}{l|}{Prompt-A-Video (DPO-1)}                  &  \underline{2.900} &  \underline{3.001} & 3.146 &  \underline{3.238} &  \underline{2.866}  & \multicolumn{1}{c|}{ \underline{3.030}}                     &  0.948  & 0.960 &   \underline{0.627} &  \underline{0.662} &  0.982 & 0.52                  \\ 
                \multicolumn{1}{l|}{Prompt-A-Video (DPO-2)}                  &  \textbf{2.930} &  \textbf{3.019} & \underline{3.183} &  \textbf{3.259} &  \textbf{2.888}  & \multicolumn{1}{c|}{ \textbf{3.056}}                     &  \underline{0.953}  & 0.959 &   \textbf{0.639} &  \textbf{0.687} &  \underline{0.983} & 0.54                  \\ \bottomrule[1pt]
		\end{tabular}
	\end{adjustbox}
	\caption{Performance comparison of different prompt boost methods on text-to-video datasets. The best scores are in \textbf{bold} and the second best scores are \underline{underlined}. Note: Abbreviations are used as follows, VQ: Visual Quality, TC: Temporal Consistency, DD: Dynamic Degree, TVA: Text-Video Alignment, FC: Factual Consistency, Avg: Average, SC: Subject Consistency, BC: Background Consistency, AQ: Aesthetic Quality, IQ: Image Quality, MS: Motion Smoothness. DPO-1: the first round of DPO, DPO-2: the second round of DPO.}
	\label{table:text-to-video results}
	\vspace{-0.3cm}
\end{table*}

Table \ref{table:text-to-video results} presents a comparison of diverse prompt boost methods within both in-domain and out-of-domain test sets. The prompts refined through our proposed methodology yield comprehensive enhancements across multiple metrics. In the WebVid test set, GPT-4o and GLM-4 fail to exhibit significant improvements over the original prompts. In contrast, our refined prompts (after DPO-2) bring about average improvements of 0.201 and 0.067 for Open-Sora 1.2 and CogVideoX, respectively. During the evaluation of the VBench test set, the metrics related to subject consistency, background consistency, and motion smoothness display minimal variations. For Open-Sora, our method increases the aesthetic quality and image quality by 0.036 and 0.028. Concerning these two metrics in CogVideoX, our method surpasses GLM-4 with more substantial enhancements (+0.094 and +0.051 as opposed to +0.069 and +0.011 of GLM-4). These results demonstrate that our refined prompts are more in line with the model's preferences, effectively bolstering the overall quality of the generated videos. 
A comparison of the last three rows of each model corroborates the finding that preference learning plays a crucial role in prompt enhancement.

Apart from our proposed method and GPT models, we also directly utilize Promptist, a text-to-image prompt enhancement model, for video generation. We notice a decline in performance within the WebVid test set and only limited improvements in VBench, indicating that image-specific modifiers do not directly translate to enhanced video quality. Notably, our refined prompts exhibit lower dynamic degree scores in VBench, potentially because the optimization process for other metrics (such as visual quality, temporal consistency) might suppress the motion magnitude. However, based on our observations and the DD results in WebVid, this reduction in motion intensity can be negligible.

\begin{figure}[htp]
	\centering
	\includegraphics[width=\linewidth]{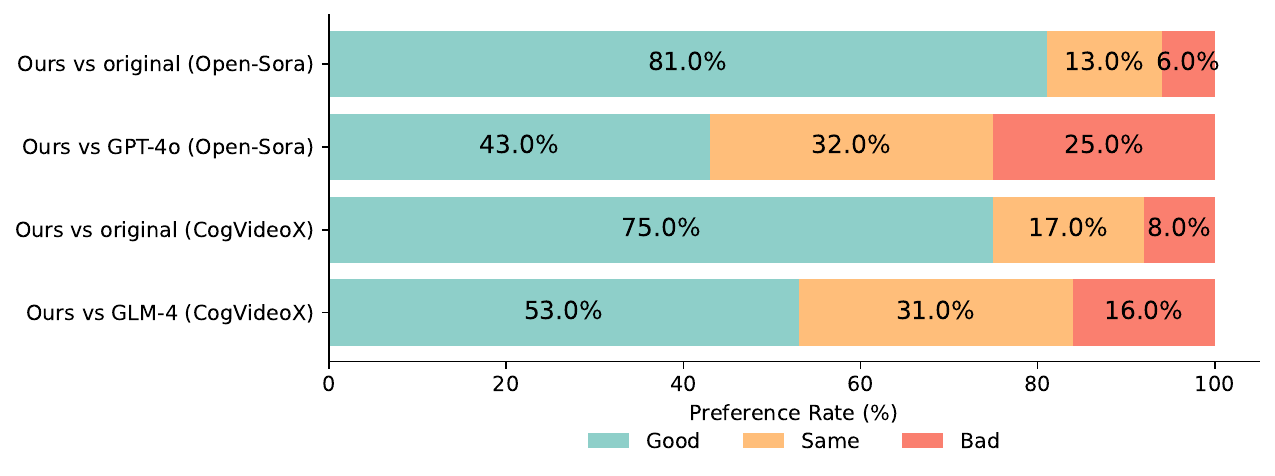}
	\caption{Win Rate of Prompt-A-Video versus original prompts and prompts refined by GPT-4o or GLM-4 with human evaluation. The user study is conducted on 100 prompts of VBench test set.}
	\label{fig:user}
    \vspace{-0.2cm}
\end{figure}

\noindent\textbf{User Study.} 
To conduct a qualitative comparison of the video generation results among different prompt enhancement methods, a user study is carried out on the VBench dataset for both Open-Sora 1.2 and CogVideoX.
We contrast the videos generated by Prompt-A-Video with those produced using original prompts and GPT-generated prompts.
Three users are tasked with comparing pairs of videos and determining which one is of superior quality.
Figure \ref{fig:user} presents a summary of the win/draw/loss rates of Prompt-A-Video. The superior win rates achieved by our method demonstrate that our prompt enhancement approach performs optimally under human subjective evaluation.

\begin{figure*}[htp]
	\centering
	\includegraphics[width=\linewidth]{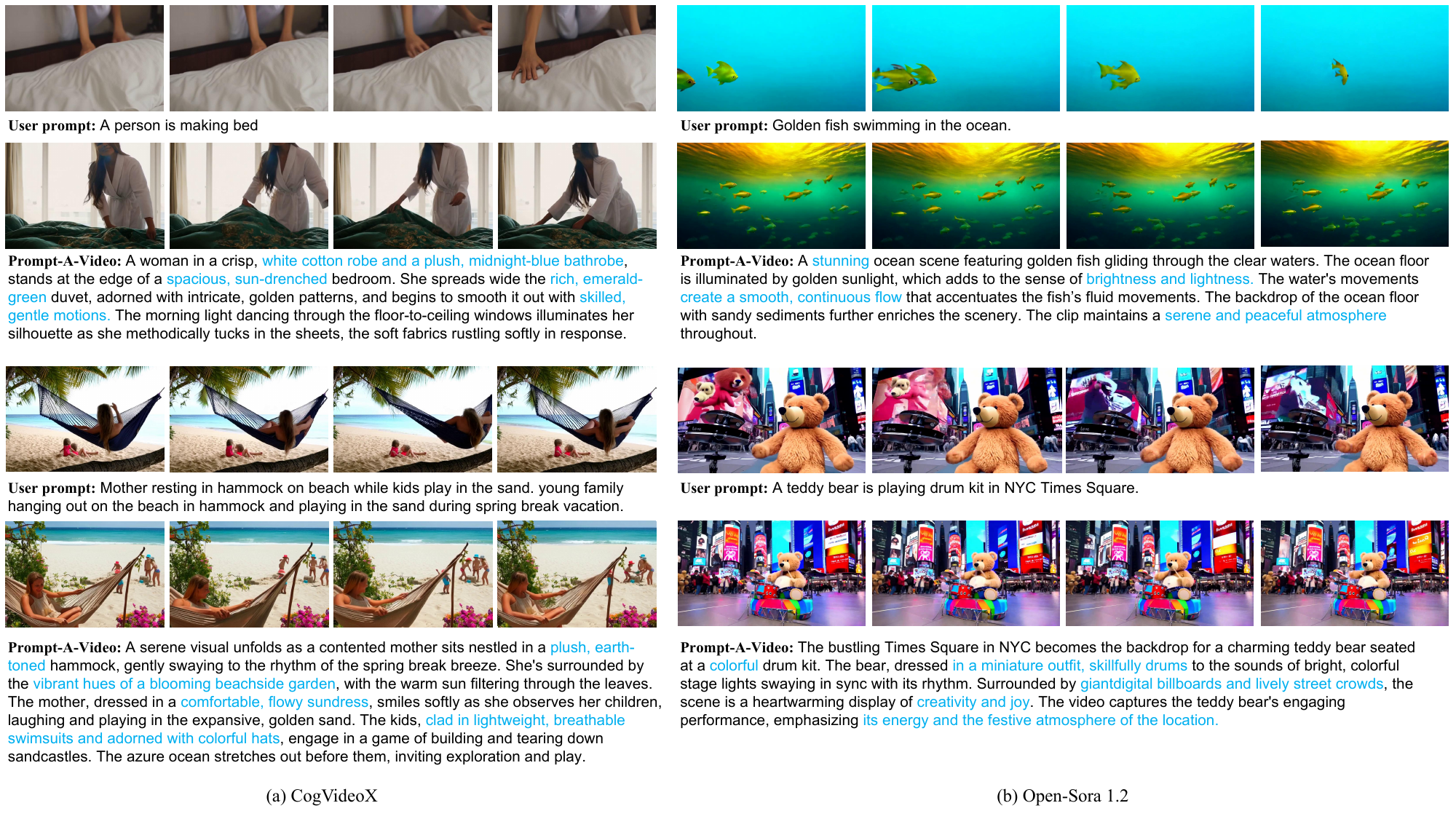}
	\caption{Videos generated using CogVideoX and Open-Sora 1.2 with user prompts and Prompt-A-Video.}
	\label{fig:visualizarion}
    \vspace{-0.2cm}
\end{figure*}

\subsection{Evaluation on Text-to-image}
To demonstrate the generalizability of our methodology, we further extend it to the text-to-image generation task. In accordance with existing works ~\cite{promptist,dynamic_boost}, we conduct experiments with \texttt{stable-diffusion-v1-4} and evaluate the performance on the widely-used HPSv2 ~\cite{hpsv2} benchmark. Specifically, we adopt the prompts from the DiffusionDB ~\cite{diffusiondb} and Lexica.art after preprocessing as the initial user prompts. Subsequently, we utilize our reward-guided prompt evolution to generate the prompt set for supervised fine-tuning. During the prompt evolution stage, we exploit the MPS ~\cite{MPS} reward model, and when creating the preference prompt for DPO tuning, we incorporate additional aesthetic and CLIP semantic alignment metrics.

As the results summarized in Table \ref{tab:t2i_hpsv2}, Prompt-A-Video exhibits superior performance on the HPSv2 benchmark, surpassing existing methods~\cite{promptist, dynamic_boost} specifically designed for text-to-image generation tasks. Specifically, our method enhances the generative results across all four types of prompts, attaining an average improvement of +0.36 points. Notably, for the prompts related to photorealistic generation, while previous methods display degraded performance compared to the original prompts, our method consistently achieves improvements.


\begin{table}[]
    \centering
    \tabcolsep=2pt
    \begin{tabular}{c|ccccc}
    \toprule[1pt]
        \textbf{Model} &  {Anime} & {Concept-Art} & {Painting} & {Photo} & {Average}\\
        \midrule
         SD1.4&  26.94 & 26.44 & 26.57 & 26.96 & 26.73 \\
         Promptist&  \underline{27.20} &  \underline{26.72} & \underline{26.75} & 26.73 & \underline{26.85} \\
         PAE &  27.06 & 26.56 & 26.67 & 26.61 & 26.73 \\
         Ours(SFT) &  26.97 & 26.52 & 26.66 & \underline{27.09} & 26.81 \\
         Ours &  \textbf{27.31} & \textbf{26.85} & \textbf{26.98} & \textbf{27.24} & \textbf{27.09} \\ 
    \bottomrule[1pt]

    \end{tabular}
    \caption{Performance comparison with the text-to-image prompt boost methods on HPSv2 benchmark.}
    \label{tab:t2i_hpsv2}
    \vspace{-0.3cm}
\end{table}



\subsection{Visualization}

We visualize two cases for each video generation model, as depicted in Figure \ref{fig:visualizarion}. The key modifiers within the enhanced prompts are highlighted in blue, thereby demonstrating more expressive prompt descriptions. Regarding the video results, the outputs generated with Prompt-A-Video enhanced prompts present superior quality, coherence, and aesthetic allure in comparison to the original ones.
For example, in the second instance of CogVideoX, the videos generated from the original prompt possess monotonous colors, conspicuous distortions of the mother and child in the third frame, and restricted motion dynamics. In contrast, our enhanced version yields more vibrant colors and clearly captures the mother's frontal view and movements.
Likewise, in the second example employing Open-Sora, the original video exhibits inconsistent and chaotic billboards in the background, while the version generated by Prompt-A-Video features coherent advertising displays and richer environmental details.

\subsection{Ablation Studies}

\subsubsection{Multi-dimensional Reward Signals}

\begin{table}[]
	\centering
	\begin{adjustbox}{max width=0.5\textwidth}
		\begin{tabular}{cccccccc}
			\toprule[1pt]
			\multicolumn{1}{c|}{Rewards}             & VQ           & TC           & DD & TVA &  FC    &     AQ & IQ      \\  \midrule 
                \multicolumn{1}{c|}{Aes + MPS}       &  \textcolor{gray}{2.660} &   2.592     &    \textcolor{gray}{3.086} &  \textcolor{gray}{3.080}   &   2.432    &   0.636  &  0.657  \\
                 \multicolumn{1}{c|}{VidS + Aes}       &  2.691  &   2.576     &   3.162  & 3.108    &  2.447     &  0.620   &  \textcolor{gray}{0.653}  \\ 
                  \multicolumn{1}{c|}{VidS + MPS}       & 2.667  &    \textcolor{gray}{2.561}    &   3.127  &  3.084   &    \textcolor{gray}{2.430}   &  \textcolor{gray}{0.614}   &  0.665  \\
                   \multicolumn{1}{c|}{VidS+Aes+MPS}       & 2.679 & 2.589 & 3.100 & 3.088 & 2.444  & 0.627  & 0.662 \\
                 \bottomrule[1pt]
		\end{tabular}
	\end{adjustbox}
	\caption{Ablation studies about different reward models combinations on VBench test set. VidS indicates VideoScore and Aes serves as Aesthetic Predictor. The used metrics contain VQ, TC, DD, TVA and FC of VideoScore, and AQ, IQ from VBench. \textcolor{gray}{Gray} values indicate the largest performance drop.}
	\label{table: ablation reward}
	\vspace{-0.3cm}
\end{table}

In our framework, the DPO phase primarily aims to align the LLM-enhanced prompts with the preferences of video generation models. Our multi-dimensional reward system offers a comprehensive preference evaluation of the generated videos, where different feedback models concentrate on distinct aspects of video quality.
Our reward models can be categorized into two types: image-level (encompassing the aesthetic predictor and MPS) and video-level (specifically, VideoScore). To evaluate the contribution of each reward model, we conduct ablation studies on the VBench.

As presented in Table \ref{table: ablation reward}, we systematically remove one reward model at a time and contrast the results with the configuration of full rewards, when conducting the first round of DPO. 
By comparing the first and last rows, it is evident that the removal of VideoScore rewards significantly deteriorates the visual quality, dynamic degree, and text alignment, these are precisely the aspects that VideoScore is designed to assess. Similarly, the absence of MPS feedback results in the lowest score for image quality, which is consistent with MPS's function in measuring overall image preferences.
In the third row, the elimination of the Aesthetic Predictor leads to performance degradation across multiple metrics, particularly in aesthetic quality and factual consistency.
The experimental results demonstrate that when trained with the combination of all feedback signals, Prompt-A-Video achieves more accurate preference alignment, and the enhanced prompts yield comprehensive performance across all metrics.


\subsubsection{Reward-guided Prompt Evolution}

\begin{figure}[t]
	\centering
	\includegraphics[width=\linewidth]{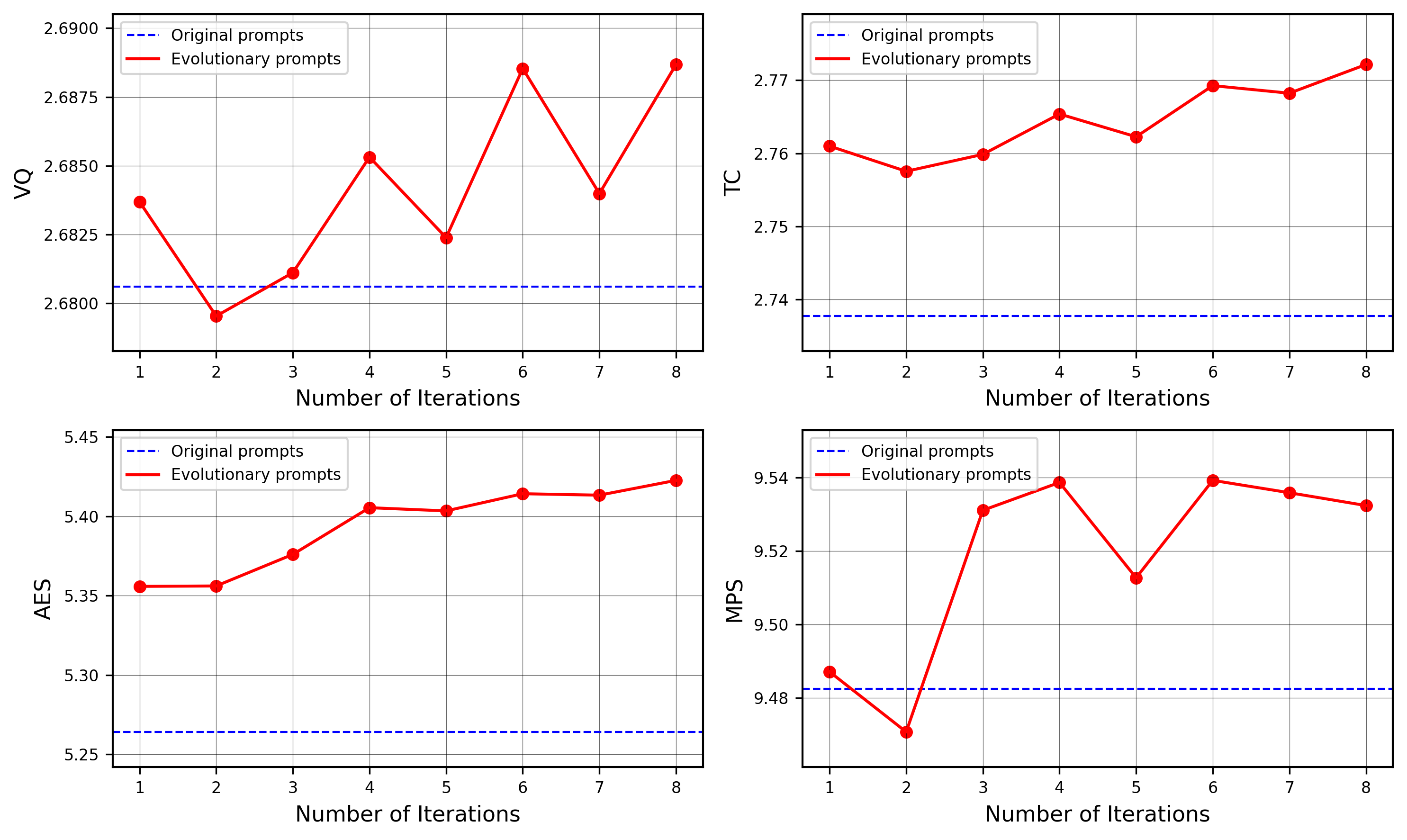}
	\caption{Video Metrics for evolutionary prompts generated in each iteration. VQ: visual quality, TC: temporal consistency, AES: aesthetic predictor, MPS: Multi-dimensional Human Preference.}
	\label{fig:statistic}
 \vspace{-0.3cm}
\end{figure}

In our reward-guided prompt evolution pipeline, GPT - 4o iteratively generates novel prompts that are better aligned with model preferences through learning from historically successful prompts. Our pipeline can effectively explore the prompt space and produce superior video quality. Figure \ref{fig:statistic} presents the prompt refinement process for Open-Sora 1.2, which illustrates the statistics of certain reward scores throughout the iterations. For each iteration, we calculate the mean value of each metric from the three newly generated prompts. The results indicate that all reward metrics display an overall upward trend across generations, thereby validating the efficacy of our evolutionary approach.


\subsubsection{Negative prompts}
Beyond prompt enhancement, we also explore approaches for negative prompt generation, which guides the model's focus towards preferred attributes and helps avoid unwanted aspects. 
While existing text-to-video models commonly employ fixed negative prompts, a question arises: Can input-specific negative prompts yield superior results by reducing particular undesired attributes associated with subjects or actions in each input?
To investigate this, we implement two adaptive negative prompt generation methods.

The first method involves leveraging the in-context learning capability of the LLM. Specifically, we manually create a small number of high-quality refined-negative prompt pairs. These pairs then serve as few-shot examples, guiding the LLM to generate adaptive negative prompts based on the input prompts. The other approach is to mimic the Prompt-A-Video posotive prompt generation process: first, fine-tune with curated prompt-negative prompt pairs, and then conduct preference alignment optimization.
Our experimental results on the VBench for CogVideoX, as shown in Table \ref{table: negative}, demonstrate that incorporating negative prompts can further enhance the performance of our videos. However, neither the in-context learning method nor the fine-tuning with preference alignment approach shows significant improvements compared to fixed negative prompts. It indicates that a comprehensive set of negative terms within fixed negative prompts is sufficient to achieve substantial performance improvements.

\begin{table}[]
	\centering
	\begin{adjustbox}{max width=0.5\textwidth}
		\begin{tabular}{ccccccc}
			\toprule[1pt]
			\multicolumn{1}{c|}{Prompts}             &  SC           & BC           & AQ & IQ &  MS & DD      \\  \midrule 
                \multicolumn{1}{c|}{Prompt-A-Video}          &  0.948  & 0.960 &  0.627 & 0.662 &  \underline{0.982} & \textbf{0.52}  \\
                 \multicolumn{1}{c|}{+ NP (fixed)}       & 0.955 & 0.964 & \textbf{0.639} & \underline{0.682} & \textbf{0.986} & 0.46   \\ 
                  \multicolumn{1}{c|}{+ NP (ICL)}       & \underline{0.958} & \underline{0.965} & \underline{0.637} & \textbf{0.683} & \textbf{0.986} & \underline{0.50}  \\
                   \multicolumn{1}{c|}{+ NP (tuning)}       & \textbf{0.959} & \textbf{0.966} & 0.634 & \underline{0.682} & \textbf{0.986} & 0.48 \\
                 \bottomrule[1pt]
		\end{tabular}
	\end{adjustbox}
	\caption{Ablation studies about different negative prompts generation methods. NP indicates negative prompts.}
	\label{table: negative}
	\vspace{-0.4cm}
\end{table}

\section{Conclusion}
Prompt-A-Video facilitates the generation of video-centric, labor-free, and preference-aligned prompts. By harnessing multidimensional rewards and evolutionary prompt refinement techniques, this framework automates the generation of high-quality prompts tailored to diverse text-to-video models. Through the integration of SFT and DPO strategies, Prompt-A-Video not only optimizes prompt quality but also ensures alignment with the inherent preferences of the video generation models. Our comprehensive evaluations demonstrate the superior performance and generalizability of our system, showcasing its efficacy in video generation and its potential for broader applications across text-to-image scenarios. 
{
    \small
    \bibliographystyle{ieeenat_fullname}
    \bibliography{main}
}

\clearpage
\setcounter{page}{1}
\maketitlesupplementary

\appendix

\section{Reward-guided Prompt Evolution}
Our reward-guided prompt evolution pipeline uses GPT-4o as evolutionary operator to automatically generate model-preferred prompts. As illustrated in Table \ref{tab: gpt instruct}, GPT-4o is guided to learn experience from historical refined prompts and generate new prompts of higher scores.

Table \ref{tab:iteration1} shows an example of prompt refinement in the first iteration. Guided by system instructions and provided examples, GPT-4o generates three different enhanced descriptions, whose scores have improvements of different degrees compared to the original prompt. 

Notably, our pipeline differs slightly from conventional evolutionary algorithms. Traditional evolutionary algorithms retain the top N results and sample K candidates from them for the next iteration based on their scores as probabilities, introducing randomness to avoid local optima. In our approach, since GPT's generation inherently possesses stochastic properties - producing varied outputs even with identical inputs - we directly advance the top N results to the next iteration.

\begin{table}[]
\begin{tabular}{p{5.8cm}p{1.4cm}}
\toprule
Attractive blonde woman doing hand massage in a spa center

& \small VQ: 2.47, TC: 2.66, DD: 2.84, TVA: 2.77, FC: 2.48, AES: 3.34, MPS: 2.7  \\ \midrule
\small
A serene scene in a spa center where an attractive blonde woman is performing a hand massage. The woman has a focused expression and is working gently. The surroundings are tranquil, with soft lighting and calming decor. The environment suggests a soothing and relaxing experience. The video does not contain any text or drastic actions.                                                                        & \small VQ: 2.63, TC: 2.73, DD: 2.92, TVA: 2.95, FC: 2.42, AES: 3.49, MPS: 3.67 \\
\small
A close-up of a blonde woman giving a hand massage in a quiet spa center. The woman's hands are applying gentle pressure, with a serene and focused expression on her face. The spa environment features soft lighting, adding to the calm and relaxing atmosphere. The scene is peaceful and intimate, designed to convey comfort and care. No text or significant movements are present in the video.                  & \small VQ: 2.58, TC: 2.77, DD: 2.88, TVA: 2.98, FC: 2.56, AES: 3.47, MPS: 3.04 \\
\small
An intimate view of a serene spa center with a blonde woman performing a hand massage. The woman appears focused and gentle, with the peaceful surroundings enhancing the calming effect. Soft, ambient lighting highlights the tranquility of the spa. The overall ambiance is relaxing, aiming to provide a sense of comfort and well-being. The video maintains a focus on the woman and her actions without any text & \small VQ: 2.58, TC: 2.69, DD: 2.77, TVA: 2.88, FC: 2.47, AES: 3.61, MPS: 3.43 \\ \bottomrule
\end{tabular}
\caption{An example of the first iteration. The scores are normalized to 0 - 5.}
\label{tab:iteration1}
\end{table}

\begin{table*}[t]\centering
\begin{minipage}{1.0\textwidth}\vspace{0mm}    \centering
\begin{tcolorbox} 
    \centering
    \begin{tabular}{p{0.97\textwidth} c}

You need to refine user's input prompt. The user's input prompt is used for video generation task. You need to refine the user's prompt to make it more suitable for the task. Here are some examples of refined prompts:\\
a close-up shot of a woman standing in a room with a white wall and a plant on the left side. the woman has curly hair and is wearing a green tank top. she is looking to the side with a neutral expression on her face. the lighting in the room is soft and appears to be natural, coming from the left side of the frame. the focus is on the woman, with the background being out of focus. there are no texts or other objects in the video. the style of the video is a simple, candid portrait with a shallow depth of field.\\
a serene scene of a pond filled with water lilies. the water is a deep blue, providing a striking contrast to the pink and white flowers that float on its surface. the flowers, in full bloom, are the main focus of the video. they are scattered across the pond, with some closer to the camera and others further away, creating a sense of depth. the pond is surrounded by lush greenery, adding a touch of nature to the scene. the video is taken from a low angle, looking up at the flowers, which gives a unique perspective and emphasizes their beauty. the overall composition of the video suggests a peaceful and tranquil setting, likely a garden or a park.\\
a serene scene in a park. the sun is shining brightly, casting a warm glow on the lush green trees and the grassy field. the camera is positioned low, looking up at the towering trees, which are the main focus of the image. the trees are dense and full of leaves, creating a canopy of green that fills the frame. the sunlight filters through the leaves, creating a beautiful pattern of light and shadow on the ground. the overall atmosphere of the video is peaceful and tranquil, evoking a sense of calm and relaxation.\\
a scene where a person is examining a dog. the person is wearing a blue shirt with the word "volunteer" printed on it. the dog is lying on its side, and the person is using a stethoscope to listen to the dog's heartbeat. the dog appears to be a golden retriever and is looking directly at the camera. the background is blurred, but it seems to be an indoor setting with a white wall. the person's focus is on the dog, and they seem to be checking its health. the dog's expression is calm, and it seems to be comfortable with the person's touch. the overall atmosphere of the video is calm and professional.\\
...\\

The refined prompt should pay attention to all objects in the video. The description should be useful for AI to re-generate the video. The description should be no more than six sentences. The refined prompt should be in English.\\
User will provide an original prompt and your revised prompts, with their generated videos' scores (Visual Quality, Temporal Consistency, Dynamic Degree, Text Video Alignment, Factual Consistency, Aesthetic score, Image quality, 7 dimensions termed as VQ, TC, DD, TVA, FC, AES, MPS), and you need to give an improved prompt according to previous prompts and their scores on different dimensions. \\
Each prompt is tagged with an index, and the sentence labeled as 0 is the initial prompt. Each prompt is followed by (VQ, TC, DD, TVA, FC, AES, MPS) scores. You need build upon the most successful prompts and learn from the high-scoring prompts. You need to observe the scores of each prompt in different aspects, learn from the experiences of previous prompts, and combine their strengths to generate better prompts.\\
The new prompts should keep the same semantic meaning with original prompt, should not add extra scene changing or too many actions, which is hard for video generation. \\
Generate 3 paraphrases of the initial prompt which keep the semantic meaning and that have higher scores than all the prompts above. Respond with each new prompt in between \textless PROMPT\textgreater and \textless/PROMPT\textgreater, e.g., \textless PROMPT\textgreater paraphrase 1\textless /PROMPT\textgreater.

    \end{tabular}
\end{tcolorbox}
    
\end{minipage}
\caption{The evolution instruction prompt for GPT-4o.}
\label{tab: gpt instruct}
\end{table*}

\section{Training Settings}

For model training, we construct around 3,000 prompt pairs for each video generation model, and use them to fine-tune Llama3-Instruct in a chat format with LoRA. The instruction prompt for LLama3-Instruct is shown in Table \ref{tab: llm instruct}. The supervised fine-tuning is deployed for 14 epochs, with a batch size of 16 and a learning rate of 1e-4. During the DPO phase, approximately 2,000 prompt triples are used for preference optimization. This stage is trained for 3 epochs with a batch size of 32 and a learning rate of 5e-5.

\begin{table*}[t]\centering
\begin{minipage}{1.0\textwidth}\vspace{0mm}    \centering
\begin{tcolorbox} 
    \centering
    \begin{tabular}{p{0.97\textwidth} c}

"dialog": 
[\\
\setlength{\parindent}{2em}\{\\
\setlength{\parindent}{4em}    "role": "user",\\
\setlength{\parindent}{4em}   
"content": "You need to refine user's input prompt. The user's input prompt is used for video generation task. You need to refine the user's prompt to make it more suitable for the task.
You will be prompted by people looking to create detailed, amazing videos. The way to accomplish this is to take their short prompts and make them extremely detailed and descriptive. You will only ever output a single video description per user request.
You should refactor the entire description to integrate the suggestions. Original prompt:\textbackslash n" + \textbf{original prompt} + "\textbackslash n New prompt:\textbackslash n"\\
\setlength{\parindent}{2em}\},\\
\setlength{\parindent}{2em}\{\\
\setlength{\parindent}{4em}"role": "assistant",\\
\setlength{\parindent}{4em}"content": \textbf{refined prompt} \\
\setlength{\parindent}{2em}\}\\
]\\


    \end{tabular}
\end{tcolorbox}
    
\end{minipage}
\caption{The instruction prompt for fine-tuning LLama3-Instruct.}
\label{tab: llm instruct}
\end{table*}

\section{Discussions}
\subsection{Model-agnostic prompt boost}

We combine fine-tuning data corresponding to each generation model to train LLaMA together, aiming to capture universal patterns in video prompt enhancement, so that the prompt booster could generalize to a broader range of video generation models. 
However, the suboptimal performance indicates distinct prompt preferences across different models, highlighting the importance of our preference-aligned approach.

\subsection{Negative prompts}

We try different generation methods of negative prompts. The fixed negative prompt is shown in Table \ref{tab: negative prompts}. As for adaptive negative prompts, considering our refined prompts from previous stages incorporate various positive modifiers, we can derive antonyms of these modifiers and negative descriptions of the subject or overall video characteristics to construct negative prompts. An exmaple is shown in Table \ref{tab: negative prompts}.

Notably, we structure these negative prompts as comma-separated descriptors rather than complete sentences containing the original prompt's content, as our empirical analysis reveals that subject-containing negative prompts can interfere with accurate generation.

\begin{table*}[t]\centering
\begin{minipage}{1.0\textwidth}\vspace{0mm}    \centering
\begin{tcolorbox} 
    \centering
    \begin{tabular}{p{0.97\textwidth} c}

\VarSty{ {\bf Fixed negative prompt} } & \\
The video is not of a high quality, it has a low resolution, and the audio quality is not clear. Strange motion trajectory, a poor composition and deformed video, low resolution, duplicate and ugly, strange body structure, long and strange neck, bad teeth, bad eyes, bad limbs, bad hands, rotating camera, blurry camera, shaking camera. Deformation, low-resolution, blurry, ugly, distortion.\\
\\
\VarSty{ {\bf Adaptive negative prompt example} } & \\
\textbf{refined prompt}: As the sun gently peeks through the vibrant window curtains, I sit comfortably in my plush, velvety chair, surrounded by an array of artfully arranged cosmetics. I begin by applying a lightweight, radiant foundation, seamlessly blending it into my skin with a fluffy brush. Next, I define my eyes with a rich, earthy eyeshadow, gradually building the palette with a pop of shimmering champagne on the lid. A swipe of deep, berry-stained lipstick completes my morning glow. I finish with a quick mist of hydrating toner and a light dusting of translucent powder, leaving my complexion fresh and flawless. The video captures my calm and focused routine, highlighting the beauty in the simplicity of a morning makeup ritual, as the natural light dancing through the room highlights the subtle, yet elegant enhancement of my daily beauty regimen.\\
\textbf{negative prompt}: The video is not of a high quality, it has a low resolution. Strange motion trajectory. bad hands, missing fingers. deformation, distortion. Dark and unclear, blur, ugly, watermark, static. The person has bad anatomy, bad eyes, bad teeth, long and strange neck, bad hands, text, error, ugly appearance, deformed body, poorly drawn face, long body. The makeup is poorly applied, with uneven lines and clashing colors. The lip color is tacky and over-saturated, and the toner makes the skin look dull and oily. The camera work is shaky and poorly framed, with harsh lighting that accentuates imperfections.
    \end{tabular}
\end{tcolorbox}
    
\end{minipage}
\caption{Negative prompt strategies.}
\label{tab: negative prompts}
\end{table*}

\section{More Visualization}

\begin{figure*}[htp]
	\centering
	\includegraphics[width=\linewidth]{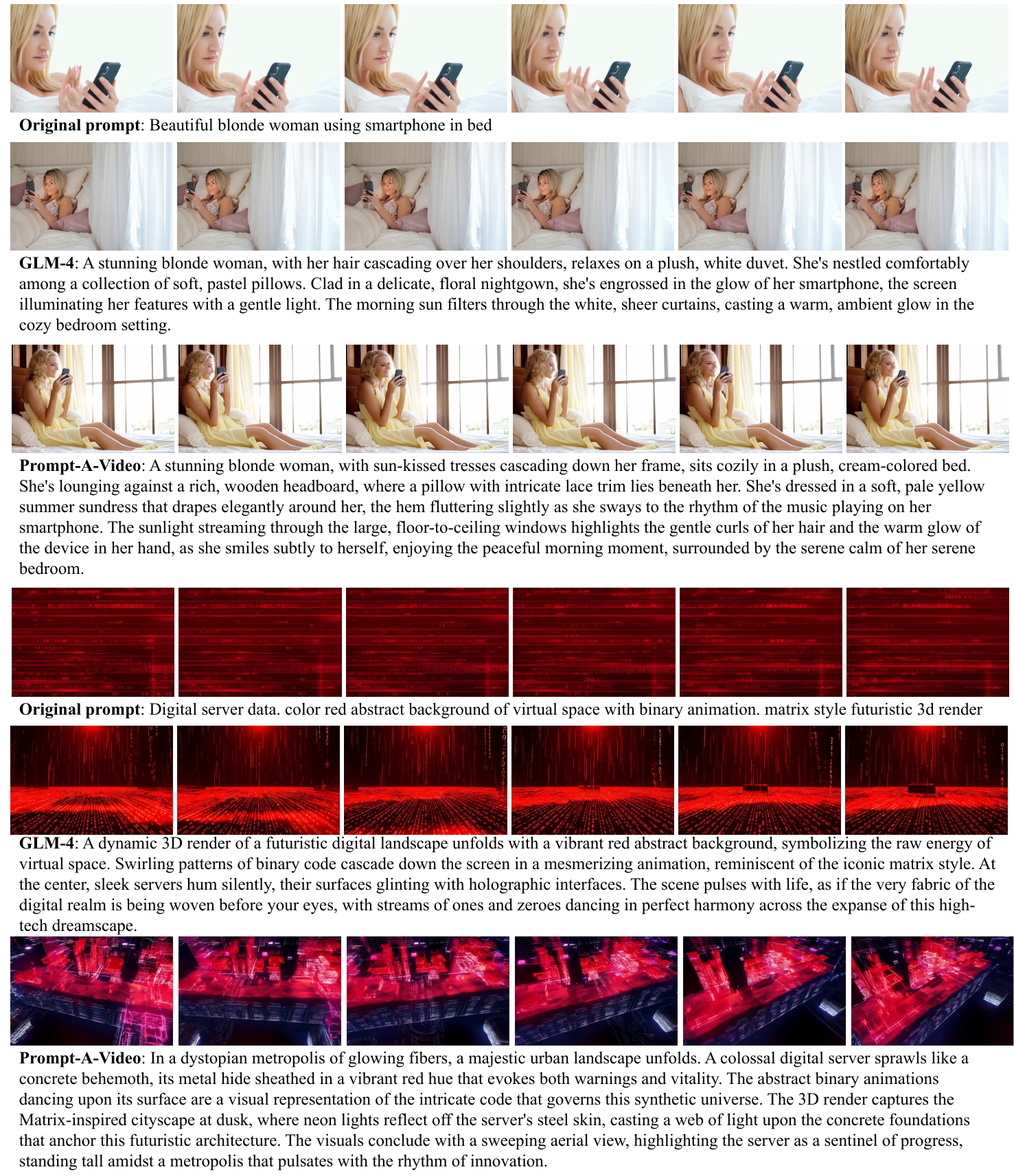}
	\caption{Videos generated using CogVideoX with user prompts, GLM4-refined prompts and Prompt-A-Video.}
	\label{fig:appendix visualizarion}
\end{figure*}

As shown in Figure \ref{fig:appendix visualizarion}, we qualitatively compare the videos generated with user prompts, GLM4-refined prompts and Prompt-A-Video. The results generated with Prompt-A-Video exhibit superior visual quality and enhanced motion magnitude.

\end{document}